\documentclass{article}

\PassOptionsToPackage{numbers,square}{natbib}
\usepackage[final]{nips_2017}

\usepackage[utf8]{inputenc} 
\usepackage[T1]{fontenc}    
\usepackage{hyperref}       
\usepackage{url}            
\usepackage{booktabs}       
\usepackage{amsfonts}       
\usepackage{nicefrac}       
\usepackage{microtype}      
\usepackage{xcolor}
\usepackage{amsmath}
\usepackage{graphicx}
\usepackage{subfiles}
\usepackage{multirow}
\usepackage{array}
\usepackage{mathbbol}
\usepackage{algorithm}
\usepackage{algorithmicx}
\usepackage{algpseudocode}

\DeclareFixedFont{\ttb}{T1}{txtt}{bx}{n}{12} 
\DeclareFixedFont{\ttm}{T1}{txtt}{m}{n}{12}  

\usepackage{color}
\definecolor{deepblue}{rgb}{0,0,0.5}
\definecolor{deepred}{rgb}{0.6,0,0}
\definecolor{deepgreen}{rgb}{0,0.5,0}

\usepackage{listings}

\newcommand\pythonstyle{\lstset{
language=Python,
basicstyle=\ttm,
otherkeywords={self},             
deletendkeywords={range},
keywordstyle=\ttb\color{deepblue},
emph={MyClass,__init__},          
emphstyle=\ttb\color{deepred},    
stringstyle=\color{deepgreen},
frame=,                         
showstringspaces=false            %
}}

\lstnewenvironment{python}[1][]
{
\pythonstyle
\lstset{#1}
}
{}

\newcommand\pythoninline[1]{{\pythonstyle\lstinline!#1!}}

\newcommand\blfootnote[1]{%
  \begingroup
  \renewcommand\thefootnote{}\footnote{#1}%
  \addtocounter{footnote}{-1}%
  \endgroup
}

\newcommand\rname{block}

\newcommand\ceil[1]{\lceil#1\rceil}

\newcommand\gradd[1]{\overline{#1\raisebox{1.9mm}{}}}
\newcommand*\samethanks[1][\value{footnote}]{\footnotemark[#1]}
\newcommand{\deriv}{\mathrm{d}}
\newcommand{\costDeriv}[1]{\overline{#1}}
\newcommand{\transpose}{\top}

\newcommand{\computationGraph}{\mathcal{G}}
\newcommand{\graphSize}{K}
\newcommand{\cost}{\mathcal{C}}
\newcommand{\nodeIdx}{i}
\newcommand{\nodeFunctionK}[1]{f_{#1}}
\newcommand{\children}{\mathrm{Child}}
\newcommand{\childIdx}{j}
\newcommand{\numConnections}{N}
\newcommand{\sequenceLength}{T}
\newcommand{\numLayers}{L}

\title{The Reversible Residual Network:\\Backpropagation Without Storing Activations}

\author{
Aidan N. Gomez\thanks{These authors contributed equally.}${\ \ }^1$,
Mengye Ren\samethanks[1]${\ \ }^{1,2}$,
Raquel Urtasun${}^{1, 2}$,
Roger B. Grosse${}^1$\\
University of Toronto${}^1$\\
Uber Advanced Technologies Group${}^2$\\
\texttt{\{aidan, mren, urtasun, rgrosse\}@cs.toronto.edu}
}

\begin{document}

\maketitle
\begin{abstract}
Deep residual networks (ResNets) have significantly pushed forward the
state-of-the-art on image classification, increasing in performance as
networks grow both deeper and wider.  However, memory consumption becomes a
bottleneck, as one needs to store  the activations in order to calculate
gradients using backpropagation. We present the Reversible Residual Network
(RevNet), a variant of ResNets where each layer's activations can be
reconstructed exactly from the next layer's. Therefore, the activations for
most layers need not be stored in memory during backpropagation. We
demonstrate the effectiveness of RevNets on CIFAR-10, CIFAR-100, and
ImageNet, establishing nearly identical classification accuracy to
equally-sized ResNets, even though the activation storage requirements are
independent of depth.\blfootnote{{Code available at \url{https://github.com/renmengye/revnet-public}}} 
\end{abstract}

\section{Introduction}
Over the last five years, deep convolutional neural networks have enabled rapid performance improvements across a wide range of visual processing tasks \citep{lecun1990handwritten,imagenet,mscoco}. For the most part, the state-of-the-art networks have been growing deeper. For instance, deep residual networks (ResNets) \citep{he2016deep} are the state-of-the-art architecture for a variety of computer vision tasks \citep{lecun1990handwritten,imagenet,mscoco}. The key architectural innovation behind ResNets was the residual block, which allows information to be passed directly through, making the backpropagated error signals less prone to exploding or vanishing. This made it possible to train networks with hundreds of layers, and this vastly increased depth led to significant performance gains.

Nearly all modern neural networks are trained using backpropagation. Since backpropagation requires storing the network's activations in memory, the memory cost is proportional to the number of units in the network. Unfortunately, this means that as networks grow wider and deeper, storing the activations imposes an increasing memory burden, which has become a bottleneck for many applications \citep{wu2016high,cyclegan}. Graphics processing units (GPUs) have limited memory capacity, leading to constraints
often exceeded by state-of-the-art architectures, some of which reach over one
thousand layers \citep{he2016deep}. Training large networks may require parallelization across multiple GPUs \citep{dean2012large,simonyan2014very}, which is both expensive and complicated to implement. Due to memory constraints, modern architectures are often trained with a mini-batch size of 1 (e.g.~\citep{wu2016high,cyclegan}), which is inefficient for stochastic gradient methods. Reducing the memory cost of storing activations would significantly improve our ability to efficiently train wider and deeper networks.

We present Reversible Residual Networks (RevNets), a variant of ResNets which is reversible in the sense that each layer's activations can be computed from the next layer's activations. This enables us to perform backpropagation without storing the activations in memory, with the exception of a handful of non-reversible layers. The result is a network architecture whose activation storage requirements are independent of depth, and typically at least an order of magnitude smaller compared with equally sized ResNets. Surprisingly, constraining the architecture to be reversible incurs no noticeable loss in performance: in our experiments, RevNets achieved nearly identical classification accuracy to standard ResNets on CIFAR-10, CIFAR-100, and ImageNet, with only a modest increase in the training time.

\section{Background}
\subsection{Backpropagation}
\label{sec:backprop}

Backpropagation \cite{rumelhart1986learning} is a classic algorithm for
computing the gradient of a cost function with respect to the parameters of a
neural network. It is used in nearly all neural network algorithms, and is
now taken for granted in light of neural network frameworks which implement
automatic differentiation \cite{abadi2016tensorflow,al2016theano}. Because
achieving the memory savings of our method requires manual implementation of
part of the backprop computations, we briefly review the algorithm.

We treat backprop as an instance of reverse mode automatic differentiation
\citep{autodiff1981}. Let $v_1, \ldots, v_\graphSize$ denote a topological
ordering of the nodes in the network's computation graph $\computationGraph$,
where $v_\graphSize$ denotes the cost function $\cost$. Each node is defined as a function
$\nodeFunctionK{\nodeIdx}$ of its parents in $\computationGraph$. Backprop
computes the total derivative $\deriv \cost / \deriv v_\nodeIdx$ for each node
in the computation graph. This total derivative defines the the effect on
$\cost$ of an infinitesimal change to $v_i$, taking into account the indirect
effects through the descendants of $v_k$ in the computation graph. Note that
the total derivative is distinct from the partial derivative $\partial f /
\partial x_i$ of a function $f$ with respect to one of its arguments $x_i$,
which does not take into account the effect of changes to $x_i$ on the other
arguments. To avoid using a small typographical difference to represent a
significant conceptual difference, we will denote total derivatives using
$\costDeriv{v_\nodeIdx} = \deriv \cost / \deriv v_\nodeIdx$.

Backprop iterates over the nodes in the computation graph in reverse
topological order. For each node $v_\nodeIdx$, it computes the total derivative
$\costDeriv{v_\nodeIdx}$ using the following rule:
\begin{equation}
\costDeriv{v_\nodeIdx} = \sum_{\childIdx \in \children(\nodeIdx)} \left( 
  \frac{\partial f_{\childIdx}}{\partial v_\nodeIdx} \right)^\transpose 
  \costDeriv{v_\childIdx}, \label{eqn:backprop}
\end{equation}
where $\children(\nodeIdx)$ denotes the children of node $v_\nodeIdx$ in
$\computationGraph$ and $\partial f_{\childIdx} / \partial v_{\nodeIdx}$
denotes the Jacobian matrix.

\subsection{Deep Residual Networks}

\begin{figure}[t]
  \centering
  \hspace{2cm}
  \includegraphics[scale=0.50]{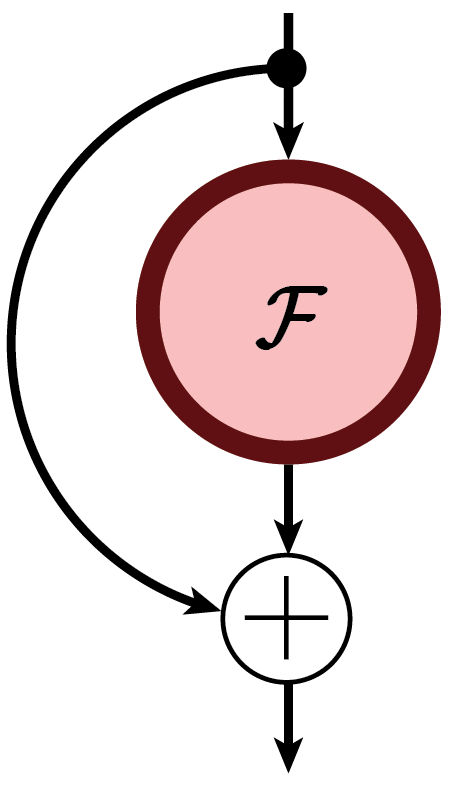}
  \hspace{1cm}
  \includegraphics[scale=0.50]{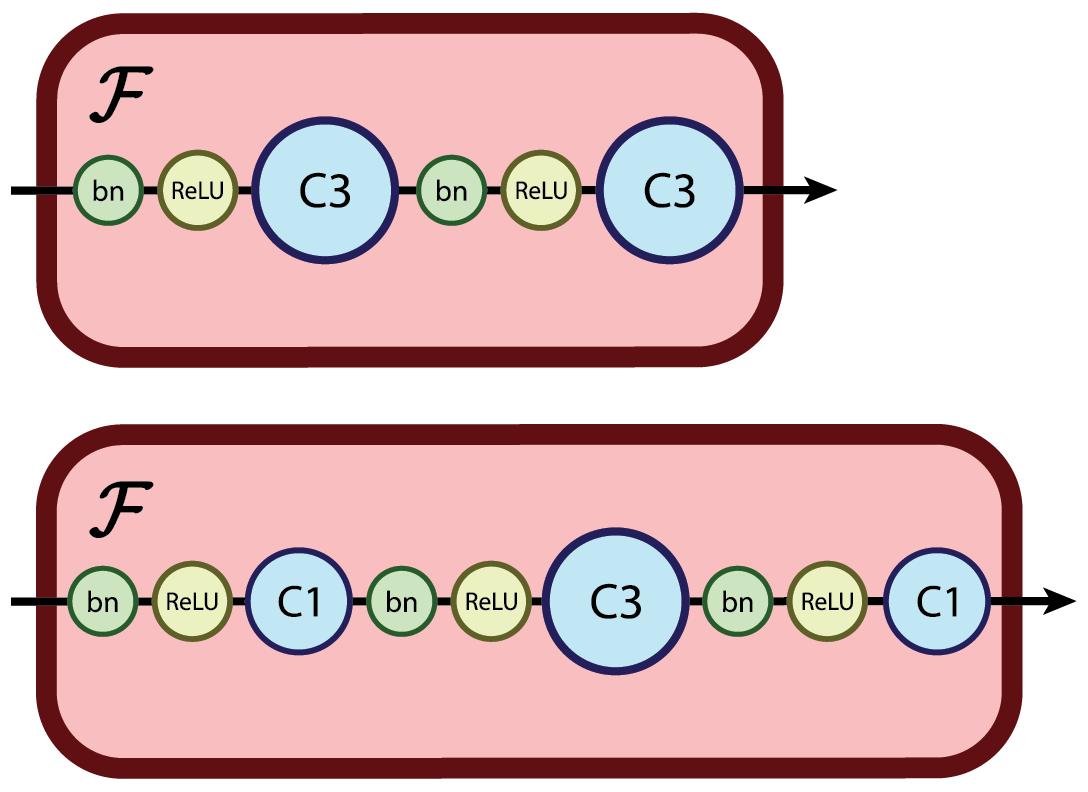}
  \caption{(left) A traditional residual \rname{} as in Equation
  \ref{residualeqn}. (right-top) A basic residual function. (right-bottom) A
  bottleneck residual function.}
  \label{resdiag}
\end{figure}

One of the main difficulties in training very deep networks is the problem of
exploding and vanishing gradients, first observed in the context of recurrent
neural networks \cite{bengio1994learning}. In particular, because a deep
network is a composition of many nonlinear functions, the dependencies across
distant layers can be highly complex, making the gradient computations
unstable. Highway networks \cite{srivastava2015highway} circumvented this
problem by introducing skip connections.  Similarly, deep residual networks
(ResNets) \cite{he2016deep} use a functional form which allows information to
pass directly through the network, thereby keeping the computations stable.
ResNets currently represent the state-of-the-art in object recognition
\citep{he2016deep}, semantic segmentation \citep{wu2016wider} and image generation \citep{pixelcnn2016}. Outside of vision, residuals have displayed impressive performance in audio generation \citep{van2016wavenet} and neural machine translation \citep{bytenet2016},

ResNets are built out of modules called residual blocks, which have the following form:

\begin{equation} \label{residualeqn}
    y = x + \mathcal{F}(x),
\end{equation}

where \(\mathcal{F}\), a function called the residual function, is typically a shallow neural net.
ResNets are robust to exploding and vanishing gradients because each residual block is able to pass signals directly through, allowing the signals to be propagated faithfully across many layers.
As displayed in Figure \ref{resdiag}, 
residual functions for image recognition generally consist of stacked batch
normalization ("BN") \citep{ioffe2015batch}, rectified linear activation
("ReLU") \citep{nair2010rectified} and convolution layers (with filters of
shape three "C3" and one "C1").

As in \citet{he2016deep}, we use two residual \rname{} architectures: the basic residual function (Figure \ref{resdiag} right-top) and the bottleneck residual function (Figure \ref{resdiag} right-bottom). The bottleneck residual consists of three convolutions, the first is a point-wise convolution which reduces the dimensionality of the feature dimension, the second is a standard convolution with filter size 3, and the final point-wise convolution projects into the desired output feature depth.

\begin{align}\label{resunits}
\begin{split}
a(x) &= ReLU(BN(x)))\\
c_k(x) &= Conv_{k \times k}(a(x))\\
\\
Basic(x) &= c_3(c_3(x))\\
Bottleneck(x) &= c_1(c_3(c_1(x)))
\end{split}
\end{align}

\subsection{Reversible Architectures}

Various reversible neural net architectures have been proposed, though for motivations distinct from our own.
\citet{maclaurin2015gradient} made use of the reversible nature of
stochastic gradient descent to tune hyperparameters via gradient descent. Our proposed method is inspired by nonlinear independent components estimation (NICE) \citep{dinh2014nice, dinh2016density}, an approach to unsupervised
generative modeling. NICE is based on learning a non-linear bijective transformation between the data space and
a latent space.  The architecture is composed of a series of blocks defined as follows, where $x_1$ and $x_2$ are a partition of the units in each layer:
\begin{equation} \label{nice}
\begin{split}
    y_1 &= x_1\\
    y_2 &= x_2 + \mathcal{F}(x_1)
\end{split}
\end{equation}
Because the model is invertible and its Jacobian has unit determinant, the log-likelihood and its gradients can be tractably computed. This architecture imposes some constraints on the functions the network can represent; for instance, it can only represent volume-preserving mappings. Follow-up work by
\citet{dinh2016density} addressed this limitation by introducing a
new reversible transformation:
\begin{equation} \label{rnvp}
\begin{split}
    y_1 &= x_1\\
    y_2 &= x_2 \odot \exp(\mathcal{F}(x_1)) + \mathcal{G}(x_1).
\end{split}
\end{equation}
Here, \(\odot\) represents the Hadamard or element-wise product. This
transformation has a non-unit Jacobian determinant due to multiplication by
\(\exp\left(\mathcal{F}(x_1)\right)\).

\section{Methods}
\begin{figure}[t]
  \centering
  (a)
  \includegraphics[scale=0.45]{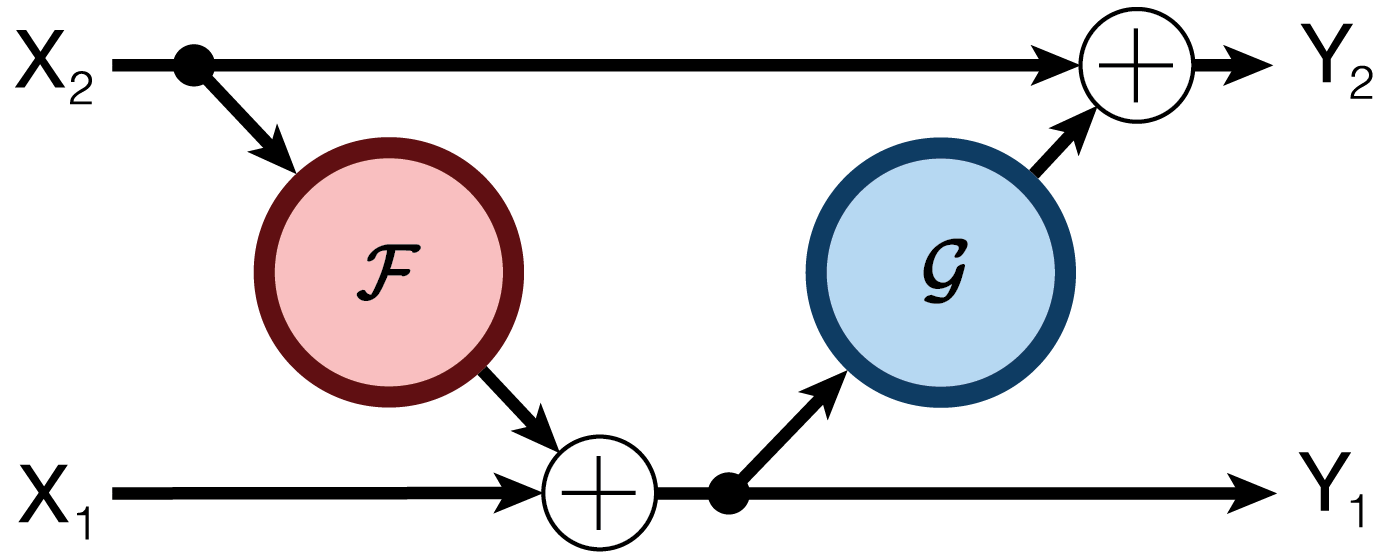}
  (b)
  \includegraphics[scale=0.45]{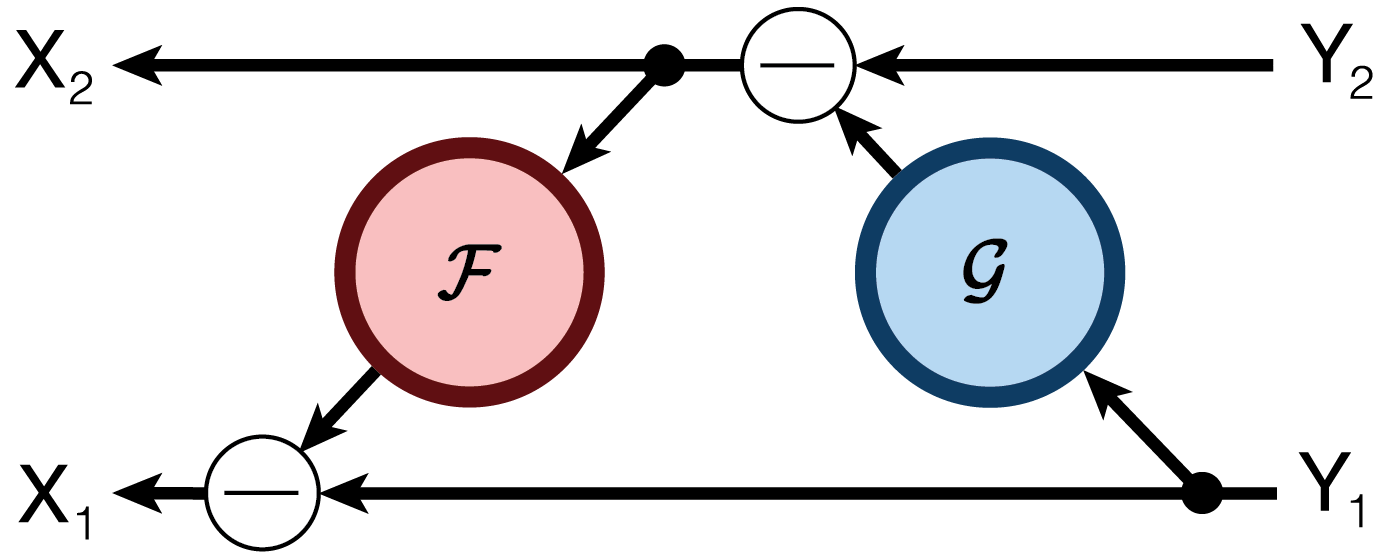}
  \caption{{\bf (a)} the forward, and {\bf (b)} the reverse computations of a residual
  \rname{}, as in Equation \ref{reversibleblock}.}
\end{figure}

We now introduce Reversible Residual Networks (RevNets), a variant of Residual Networks which is reversible in the sense that each layer's activations can be computed from the next layer's activations. We discuss how to reconstruct the activations online during backprop, eliminating the need to store the activations in memory. 

\subsection{Reversible Residual Networks}

RevNets are composed of a series of reversible blocks, which we now define. We must partition the units in each layer into two groups, denoted $x_1$ and $x_2$; for the remainder of the paper, we assume this is done by partitioning the channels, since we found this to work the best in our experiments.\footnote{The possibilities we explored included columns, checkerboard, rows and
channels, as done by \cite{dinh2016density}. We found that performance was consistently superior using the
channel-wise partitioning scheme and comparable across the remaining options. We note that channel-wise partitioning has also been explored in the context of multi-GPU
training via 'grouped' convolutions \citep{krizhevsky2012imagenet}, and more
recently, convolutional neural networks have seen significant success by way of
'separable' convolutions \citep{sifre2014rigid, chollet2016xception}. }
Each reversible block takes inputs $(x_1, x_2)$ and produces outputs $(y_1, y_2)$ according to the following additive coupling rules -- inspired by NICE's \citep{dinh2014nice}
transformation in Equation \ref{nice} -- and residual functions \(\mathcal{F}\) and 
\(\mathcal{G}\) analogous to those in standard ResNets:
\begin{align}
y_1 &= x_1 + \mathcal{F}(x_2) \nonumber \\
y_2 &= x_2 + \mathcal{G}(y_1) 
\end{align}
Each layer's activations can be reconstructed from the next layer's activations as follows:
\begin{align}
x_2 &= y_2 - \mathcal{G}(y_1) \nonumber \\
x_1 &= y_1 - \mathcal{F}(x_2)
\end{align}

Note that unlike residual blocks, reversible blocks must have a stride of 1 because otherwise the layer discards information, and therefore cannot be reversible. Standard ResNet architectures typically have a handful of layers with a larger stride. If we define a RevNet architecture analogously, the activations must be stored explicitly for all non-reversible layers.

\subsection{Backpropagation Without Storing Activations}\label{gradientcomp}

To derive the backprop procedure, it is helpful to
rewrite the forward (left) and reverse (right) computations in the following way:
\begin{alignat}{2} \label{reversibleblock}
    z_1 &= x_1 + \mathcal{F}(x_2) \qquad&\qquad z_1 &= y_1 \nonumber\\
    y_2 &= x_2 + \mathcal{G}(z_1) \qquad&\qquad x_2 &= y_2 - \mathcal{G}(z_1)\\
    y_1 &= z_1 \qquad&\qquad x_1 &= z_1 - \mathcal{F}(x_2) \nonumber
\end{alignat}
Even though $z_1 = y_1$, the two variables represent distinct nodes of the computation
graph, so the total derivatives $\gradd{z_1}$ and $\gradd{y_1}$ are
different. In particular, $\gradd{z_1}$ includes the indirect effect
through $y_2$, while $\gradd{y_1}$ does not. This splitting lets us
implement the forward and backward passes for reversible blocks in a modular
fashion. 
In the backwards pass, we are given the activations $(y_1, y_2)$ and their total
derivatives $(\gradd{y_1}, \gradd{y_2})$ and wish to compute the
inputs $(x_1, x_2)$, their total derivatives $(\gradd{x_1},
\costDeriv{x_2})$, and the total derivatives for any parameters associated with ${\cal F}$ and ${\cal G}$. (See Section~\ref{sec:backprop} for our backprop notation.)
We do this by combining the reconstruction formulas (Eqn.~\ref{reversibleblock})
with the backprop rule (Eqn.~\ref{eqn:backprop}). The resulting algorithm is
given as Algorithm \ref{revgradd}.\footnote{We assume for notational clarity
that the residual functions do not share parameters, but Algorithm
\ref{revgradd} can be trivially extended to a network with weight sharing, such
as a recurrent neural net.}

By applying Algorithm \ref{revgradd} repeatedly, one can perform backprop on a
sequence of reversible blocks if one is given simply the activations and their
derivatives for the top layer in the sequence. In general, a practical
architecture would likely also include non-reversible layers, such as
subsampling layers; the inputs to these layers would need to be stored explicitly
during backprop. However, a typical ResNet architecture involves long sequences
of residual blocks and only a handful of subsampling layers; if we mirror the
architecture of a ResNet, there would be only a handful of non-reversible
layers, and the number would not grow with the depth of the network. In this
case, the storage cost of the activations would be small, and independent of
the depth of the network.

{\bf Computational overhead.} In general, for a network with $\numConnections$
connections, the forward and backward passes of backprop require approximately
$\numConnections$ and $2\numConnections$ add-multiply operations,
respectively. For a RevNet, the residual functions each must be recomputed
during the backward pass. Therefore, the number of operations required for
reversible backprop is approximately $4 \numConnections$, or roughly $33\%$
more than ordinary backprop. (This is the same as the overhead
introduced by checkpointing \cite{martens2012training}.) In practice, we have found the forward and backward passes to be about equally expensive on GPU architectures; if this is the case, then the computational overhead of RevNets is closer to $50\%$.

\begin{algorithm}
\caption{Reversible Residual Block Backprop}\label{revgradd}
\begin{algorithmic}[1]
\Function{BlockReverse}{$(y_1, y_2), (\gradd{y}_1,\gradd{y}_2)$}
   \State $z_1\gets y_1$
   \State $x_2\gets y_2 - {\cal G}(z_1)$
   \State $x_1\gets z_1 - {\cal F}(x_2)$
   \State $\gradd{z}_1\gets \gradd{y}_1 + 
      \left( \frac{\partial {\cal G}}{\partial z_1} \right)^\top 
      \gradd{y}_2$  \Comment{ordinary backprop}
   \State $\gradd{x}_2\gets \gradd{y}_2 + 
      \left( \frac{\partial {\cal F}}{\partial x_2} \right)^\top 
      \gradd{z}_1$ \Comment{ordinary backprop}
   \State $\gradd{x}_1\gets \gradd{z}_1$
   \State \(\gradd{w}_{\cal F}\gets 
      \left( \frac{\partial {\cal F}}{\partial w_{\cal F}} \right)^\top 
      \gradd{z}_1\) \Comment{ordinary backprop}
   \State \(\gradd{w}_{\cal G}\gets 
      \left( \frac{\partial {\cal G}}{\partial w_{\cal G}} \right)^\top 
      \gradd{y}_2\) \Comment{ordinary backprop}
   \State \textbf{return} $(x_1, x_2)$ and 
      $(\gradd{x}_1, \gradd{x}_2)$ and 
      \((\gradd{w}_{\cal F}, \gradd{w}_{\cal G})\)
\EndFunction
\end{algorithmic}
\end{algorithm}

{\bf Modularity.} Note that Algorithm \ref{revgradd} is agnostic to the form of
the residual functions ${\cal F}$ and ${\cal G}$. The steps which use the
Jacobians of these functions are implemented in terms of ordinary backprop,
which can be achieved by calling automatic differentiation routines
(e.g.~\verb+tf.gradients+ or \verb+Theano.grad+). Therefore, even though
implementing our algorithm requires some amount of manual implementation of
backprop, one does not need to modify the implementation in order to change the
residual functions.

{\bf Numerical error.} While Eqn.~\ref{reversibleblock} reconstructs the
activations exactly when done in exact arithmetic, practical \verb+float32+
implementations may accumulate numerical error during backprop. We study the
effect of numerical error in Section~\ref{rev_net_perf}; while the error is
noticeable in our experiments, it does not significantly affect final
performance. We note that if numerical error becomes a significant issue, one
could use fixed-point arithmetic on the $x$'s and $y$'s (but ordinary floating
point to compute ${\cal F}$ and ${\cal G}$), analogously to
\cite{maclaurin2015gradient}. In principle, this would enable exact
reconstruction while introducing little overhead, since the computation of the
residual functions and their derivatives (which dominate the computational
cost) would be unchanged.

\section{Related Work}
A number of steps have been taken towards reducing the storage requirements
of extremely deep neural networks. Much of this work has
focused on the modification of memory allocation within the training algorithms
themselves \citep{abadi2016tensorflow,al2016theano}.  
Checkpointing \citep{martens2012training, chen2016training,
gruslys2016memory} is one well-known technique which trades off spatial and temporal complexity; during backprop, one stores a subset of the activations (called checkpoints) and recomputes the remaining activations as required.
\citet{martens2012training} adopted this technique in the context of training
recurrent neural networks on a sequence of length \(\sequenceLength\) using backpropagation
through time \citep{williams1989learning}, storing every \(\ceil{\sqrt{\sequenceLength}}\)
layers and recomputing the intermediate activations between each
during the backward pass.  \citet{chen2016training} later proposed to
recursively apply this strategy on the sub-graph between checkpoints.
\citet{gruslys2016memory} extended this approach by applying dynamic
programming to determine a storage strategy which minimizes the computational cost for a given memory budget.

To analyze the computational and memory complexity of these alternatives, assume for simplicity a feed-forward network consisting of $\numLayers$ identical layers.
Again, for simplicity, assume the units are chosen such that the cost of forward propagation or backpropagation through a single layer is 1, and the memory cost of storing a single layer's activations is 1. In this case, ordinary backpropagation has computational cost $2\numLayers$ and storage cost $\numLayers$ for the activations.
The method of \citet{martens2012training} requres \(2\sqrt{\numLayers}\) storage, and it
demands an additional forward computation for each layer, leading to a total
computational cost of \(3\numLayers\). The recursive algorithm of \citet{chen2016training} reduces the required memory to \(\mathcal{O}(\log
\numLayers)\), while increasing the computational cost to \(\mathcal{O}(\numLayers \log \numLayers)\).
In comparison to these, our method incurs 
\(\mathcal{O}(1)\) storage cost --- as only a single \rname{} must be stored --- and
computational cost of 3$\numLayers$. The time and space complexities of these
methods are summarized in Table~\ref{tab:complexities}.

Another approach to saving memory is to replace backprop itself. The decoupled neural interface
\citep{synthgrad2016} updates each weight matrix using a gradient
approximation, termed the \emph{synthetic gradient}, computed based on only the node's activations instead of the
global network error. This removes any long-range gradient computation dependencies
in the computation graph, leading to \(\mathcal{O}(1)\) activation storage
requirements. However, these savings are achieved only after the synthetic gradient estimators have been trained; that training requires all the activations to be stored.

\begin{table}[t]
  \centering
  \caption{Computational and spatial complexity comparisons. $\numLayers$ denotes the number of layers.}\label{tab:complexities}
  \begin{tabular}{lll}
    \toprule
    \multirow{2}{*}{Technique} & Spatial Complexity & Computational \\
    & (Activations) & Complexity \\
    \midrule
    Naive & \(\mathcal{O}(\numLayers)\) & \(\mathcal{O}(\numLayers)\) \\
    Checkpointing \citep{martens2012training} & 
    \(\mathcal{O}(\sqrt{\numLayers})\) & \(\mathcal{O}(\numLayers)\) \\
    Recursive Checkpointing \citep{chen2016training} & 
    \(\mathcal{O}(\log \numLayers)\) & 
    \(\mathcal{O}(\numLayers\log \numLayers)\) \\
    Reversible Networks (Ours) & \(\mathcal{O}(1)\) & \(\mathcal{O}(\numLayers)\) \\
  \bottomrule
  \end{tabular}\\[4mm]
\end{table}

\section{Experiments}
We experimented with RevNets on three standard image classification
benchmarks: CIFAR-10, CIFAR-100, \citep{cifar10-100} and ImageNet \citep{imagenet}. 
In order to make our results directly comparable with standard ResNets, we tried
to match both the computational depth and the number of parameters as closely as possible.
We observed that each
reversible \rname{} has a computation depth of two original residual \rname{}s.
Therefore, we reduced the total number of residual \rname{}s by approximately half,
while approximately doubling the number of channels per \rname{}, since they are
partitioned into two. Table \ref{tab:arch_detail} shows the details of the
RevNets and their corresponding traditional ResNet. In all of our experiments, we were interested in whether our RevNet architectures (which are far more memory efficient) were able to match the classification accuracy of ResNets of the same size.

\subsection{Implementation}

\begin{table}[t]
  \centering
  \caption{Architectural details. 'Bottleneck' indicates whether the residual unit type used was the \(Bottleneck\) or \(Basic\) variant (see Equation \ref{resunits}). 'Units' indicates the number of residual units in each group. 'Channels' indicates the number of filters used in each unit in each group. 'Params' indicates the number of parameters, in millions, each network uses.}
  \label{tab:arch_detail}
  \begin{tabular}{c|ccccc}
  \toprule
  Dataset        & Version       & Bottleneck & Units    & Channels        & Params (M)  \\
  \midrule
  CIFAR-10 (100) & ResNet-32     & No         & 5-5-5    & 16-16-32-64     & 0.46 (0.47) \\ 
  CIFAR-10 (100) & RevNet-38  & No         & 3-3-3    & 32-32-64-112    & 0.46 (0.48) \\
  \midrule
  CIFAR-10 (100) & ResNet-110    & No         & 18-18-18 & 16-16-32-64     & 1.73 (1.73) \\ 
  CIFAR-10 (100) & RevNet-110 & No         & 9-9-9    & 32-32-64-128    & 1.73 (1.74) \\
  \midrule
  CIFAR-10 (100) & ResNet-164    & Yes        & 18-18-18 & 16-16-32-64     & 1.70 (1.73) \\ 
  CIFAR-10 (100) & RevNet-164 & Yes        & 9-9-9    & 32-32-64-128    & 1.75 (1.79) \\
  \midrule
  ImageNet       & ResNet-101    & Yes        & 3-4-23-3 & 64-128-256-512  & 44.5        \\
  ImageNet       & RevNet-104 & Yes        & 2-2-11-2 & 128-256-512-832 & 45.2        \\
  \bottomrule
  \end{tabular}
\end{table}

We implemented the RevNets using the TensorFlow library
\citep{abadi2016tensorflow}.  We manually make calls to TensorFlow's automatic
differentiation method (i.e.  \texttt{tf.gradients}) to construct the
backward-pass computation graph without referencing activations computed in the
forward pass. While building the backward graph, we reconstruct the input
activations ($\hat{x}_1, \hat{x}_2$) for each \rname{} (Equation
\ref{reversibleblock}); Second, we apply \texttt{tf.stop\_gradient} on the
reconstructed inputs to prevent auto-diff from traversing into the reconstructions'
computation graph, then call the forward functions again to compute
($\hat{y}_1, \hat{y}_2$) (Equation \ref{reversibleblock}).  Lastly, we use
auto-diff to traverse from ($\hat{y}_1, \hat{y}_2$) to ($\hat{x}_1, \hat{x}_2$)
and the parameters ($w_{\cal F}, w_{\cal G}$). This implementation leverages
the convenience of the auto-diff functionality to avoid manually deriving
gradients; however the computational cost becomes $5N$, compared with $4N$ for Algorithm \ref{revgradd}, and  $3N$ for ordinary backpropagation (see Section \ref{gradientcomp}). The full theoretical efficiency can be realized by reusing
the \(\cal F\) and \(\cal G\) graphs' activations that were computed in the
reconstruction steps (lines 3 and 4 of Algorithm \ref{revgradd}).

\begin{table}[t]
  \centering
  \caption{Classification error on CIFAR}\label{tab:cifar}
  \begin{tabular}{cccccccc}
    \toprule
    \multirow{2}{*}{\vspace{-2mm}Architecture} & 
    \multicolumn{2}{c}{CIFAR-10 \citep{cifar10-100}} && 
    \multicolumn{2}{c}{CIFAR-100 \citep{cifar10-100}} \\
    \cmidrule{2-3}\cmidrule{5-6}
     & ResNet & RevNet && ResNet & RevNet \\
    \midrule
    32 (38) & \bf{7.14\%} & 7.24\%      && 29.95\%      & \bf{28.96\%} \\
    110     & \bf{5.74\%} & 5.76\%      && 26.44\%      & \bf{25.40\%}  \\
    164     & 5.24\%      & \bf{5.17\%} && \bf{23.37\%} & 23.69\%       \\
  \bottomrule
  \end{tabular}\\[4mm]
\end{table}

\begin{table}[t]
  \centering
  \caption{Top-1 classification error on ImageNet (single crop)}\label{tab:imagenet}
  \begin{tabular}{cc}
    \toprule
    ResNet-101 & RevNet-104 \\
    \midrule
    \bf{23.01\%} & 23.10\% \\
  \bottomrule
  \end{tabular}\\[4mm]
\end{table}

\begin{figure}[t]
\centering
\small
\includegraphics[width=0.45\textwidth]{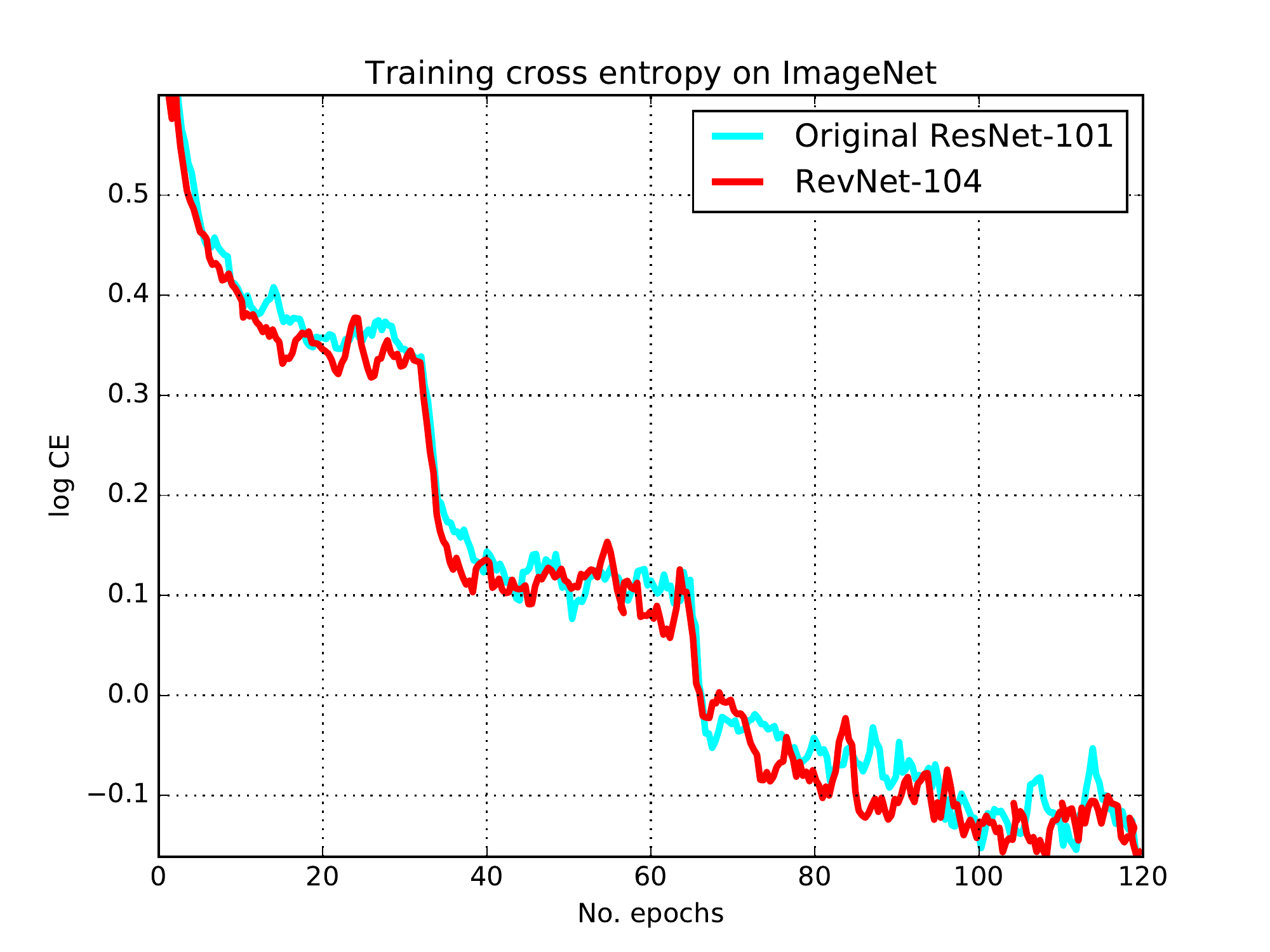}
\includegraphics[width=0.45\textwidth]{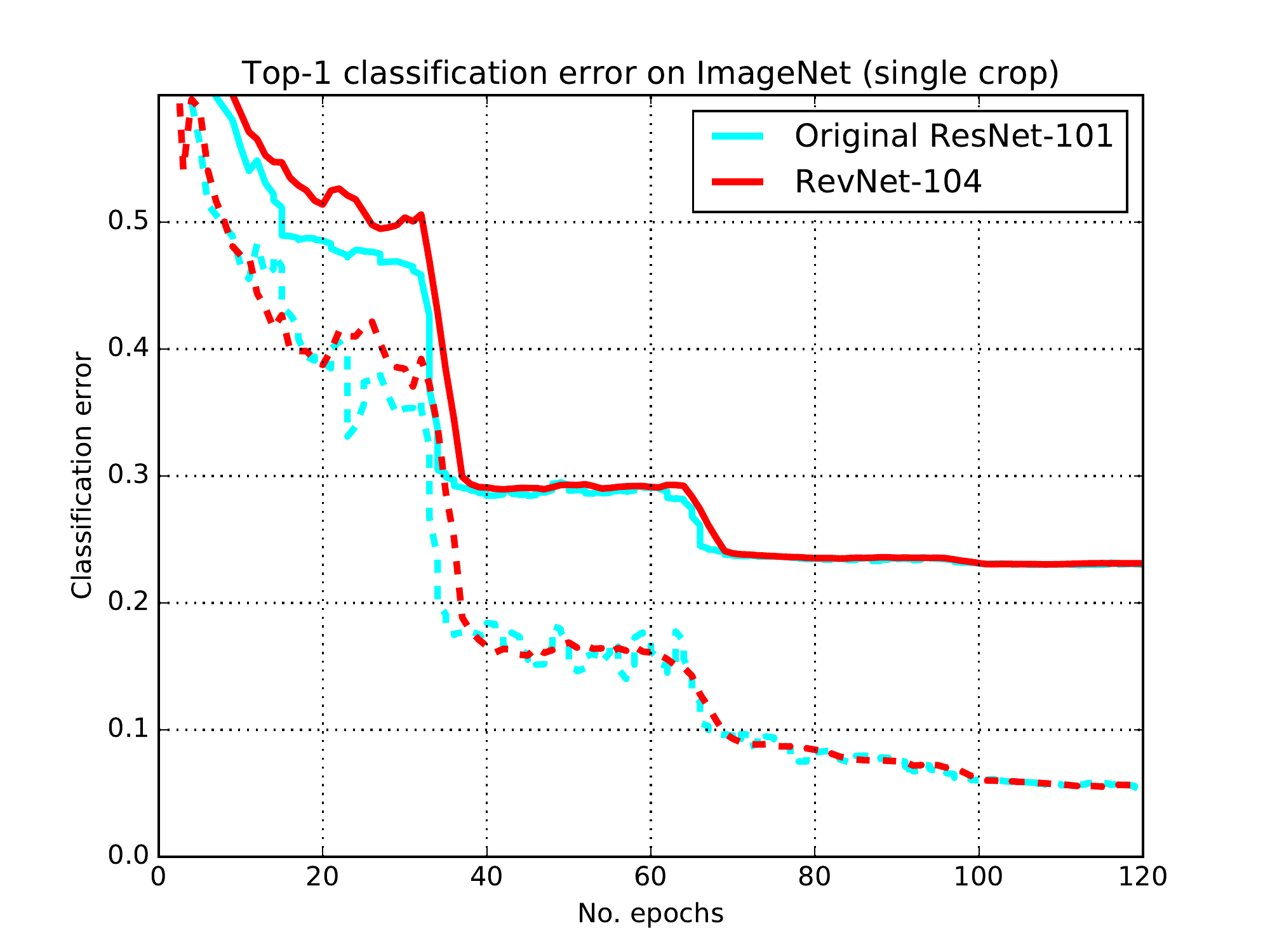}
\caption{Training curves for ResNet-101 vs.~RevNet-104 on ImageNet, with both
  networks having approximately the same depth and number of free parameters.
  {\bf Left:} training cross entropy; {\bf Right:} classification error, where
  dotted lines indicate training, and solid lines validation.}
\label{fig:imgnet}
\end{figure}

\begin{figure}[t]
\centering
\small
\includegraphics[width=0.32\textwidth]{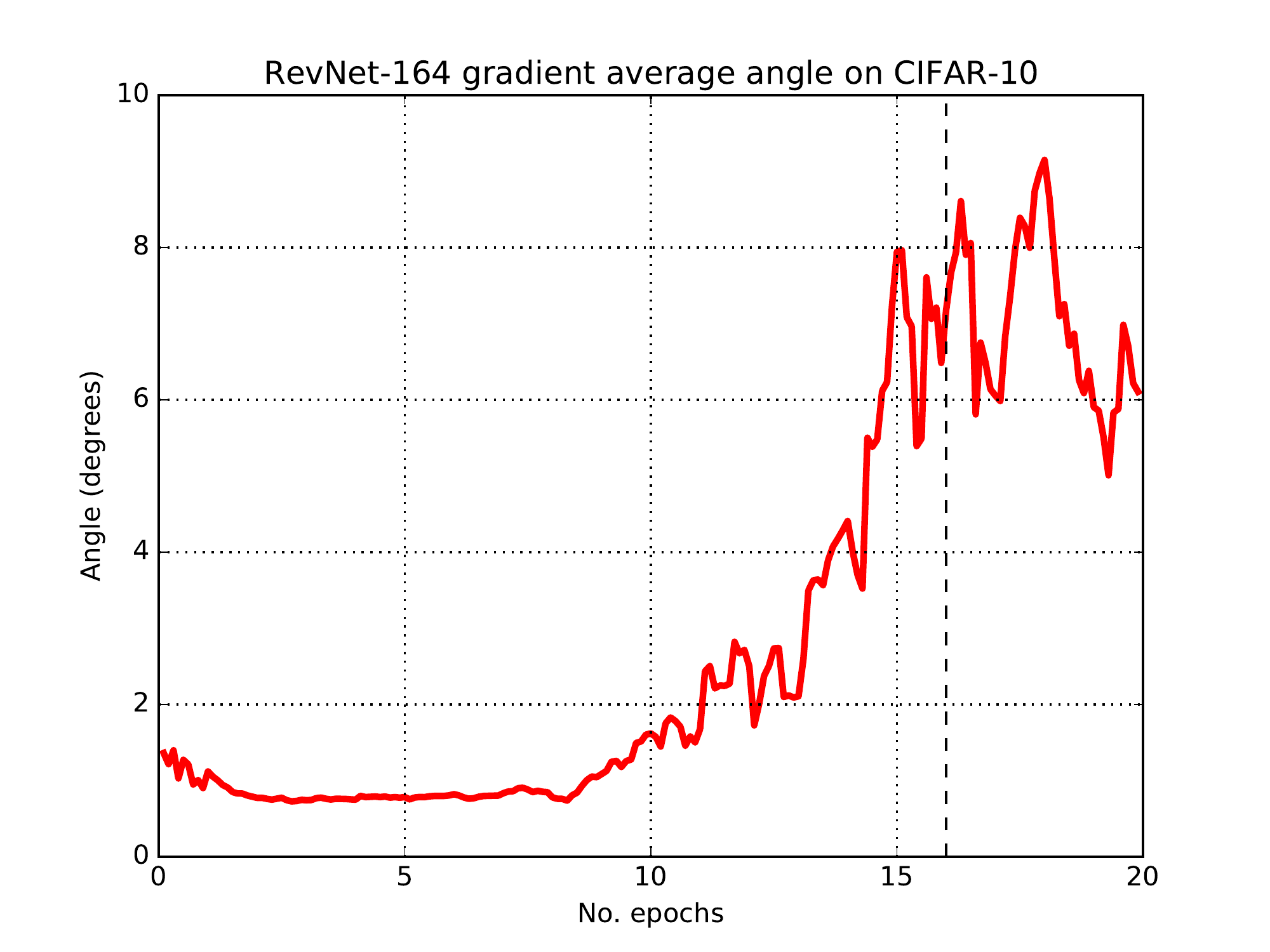}
\includegraphics[width=0.32\textwidth]{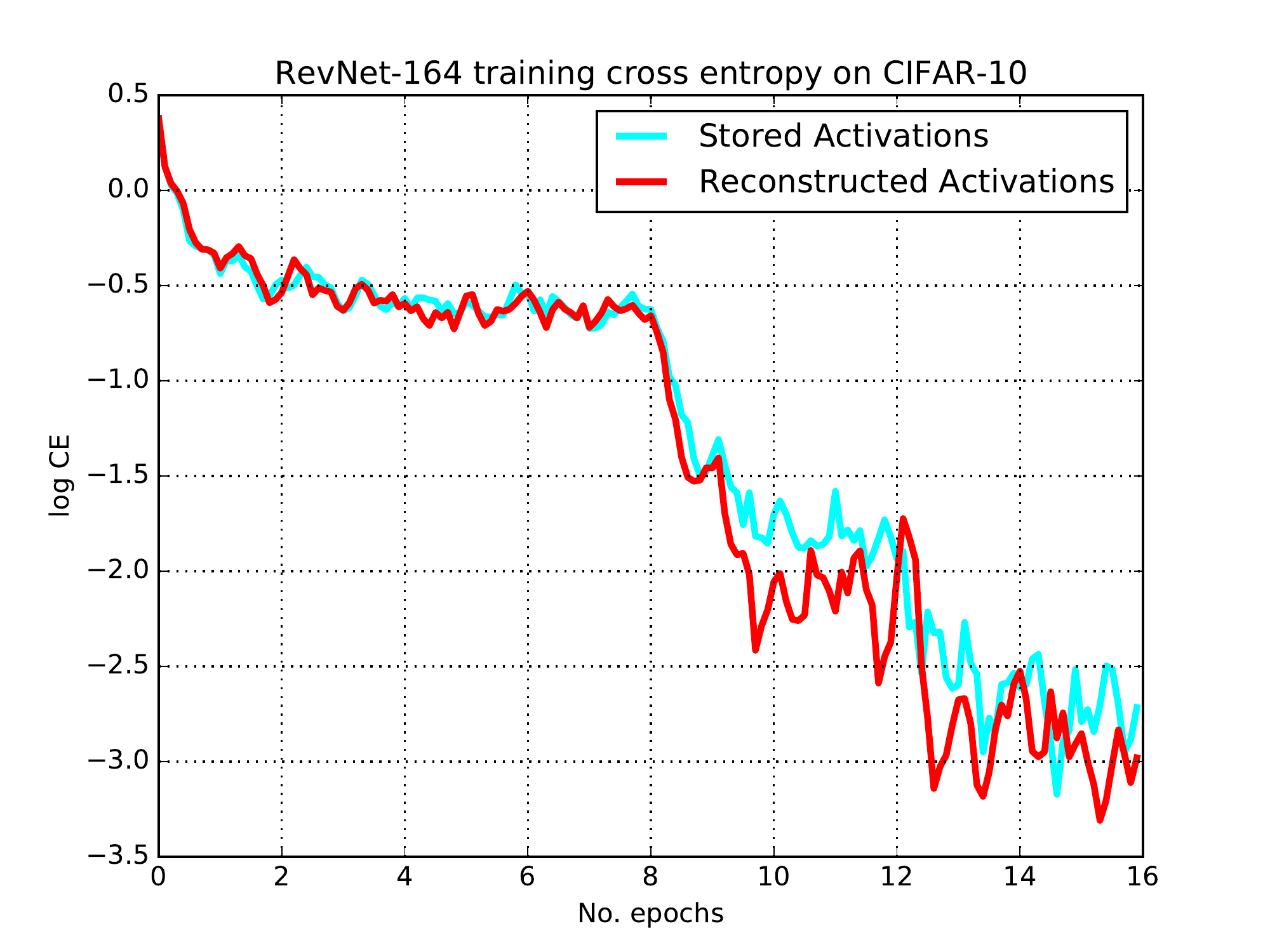}
\includegraphics[width=0.32\textwidth]{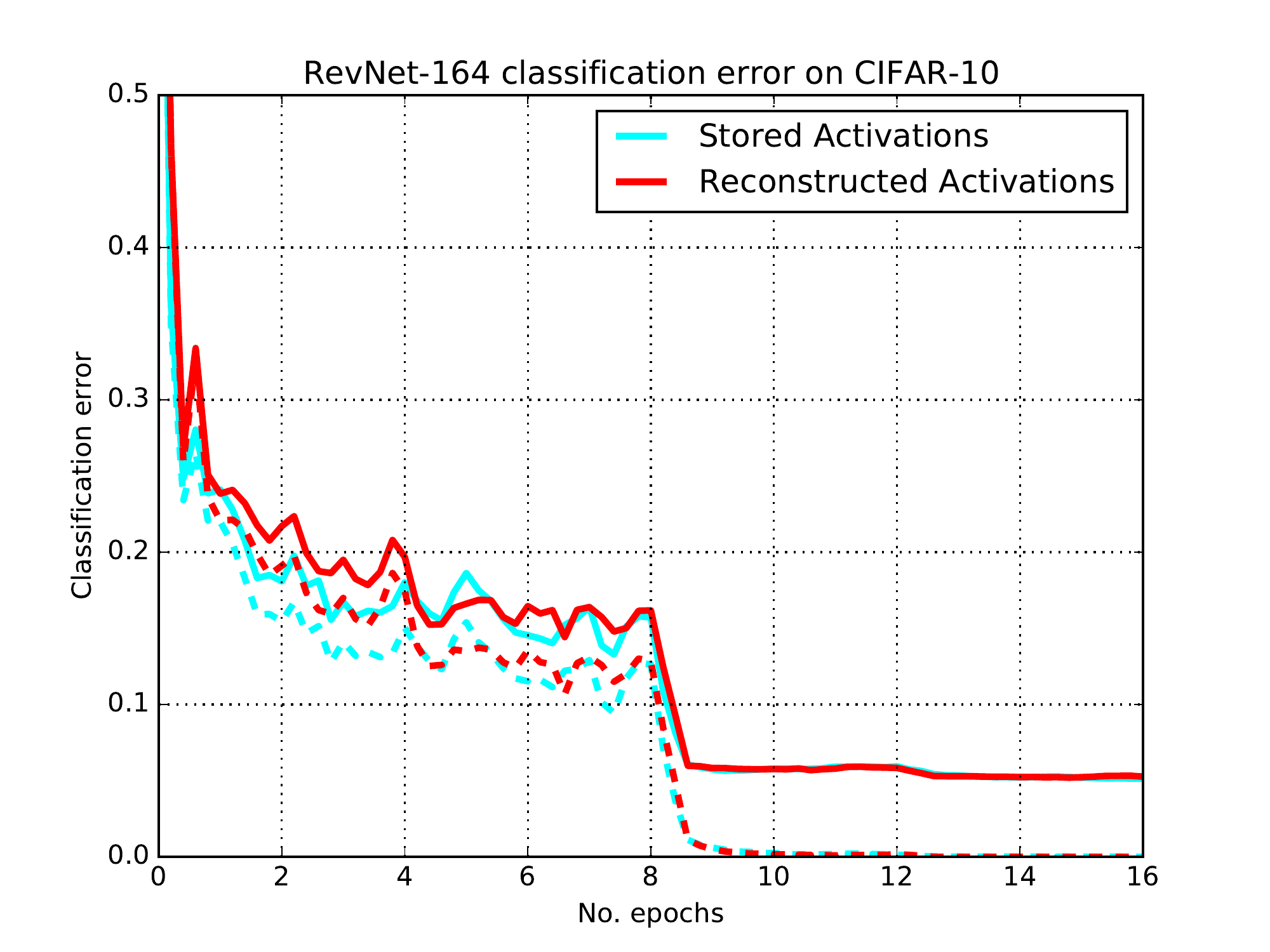}
\caption{{\bf Left:} angle (degrees) between
  the gradient computed using stored and reconstructed activations throughout
  training. While the angle grows during training, it remains small in
  magnitude. We measured 4 more epochs after regular training length and did not 
  observe any instability. 
  {\bf Middle:} training cross entropy; 
  {\bf Right:} classification
  error, where dotted lines indicate training, and solid lines validation; No
  meaningful difference in training efficiency or final performance was
  observed between stored and reconstructed activations.}
\label{fig:auto_vs_manual}
\end{figure}

\subsection{RevNet performance}
\label{rev_net_perf}
Our ResNet implementation roughly matches the previously reported
classification error rates \cite{he2016deep}. As shown in Table
\ref{tab:cifar}, our RevNets roughly matched the error rates of traditional
ResNets (of roughly equal computational depth and number of parameters) on
CIFAR-10 \& 100 as well as ImageNet (Table \ref{tab:imagenet}). In no condition
did the RevNet underperform the ResNet by more than 0.5\%, and in some cases,
RevNets achieved slightly better performance.  Furthermore, Figure
\ref{fig:imgnet} compares ImageNet training curves of the ResNet and RevNet
architectures; reversibility did not lead to any noticeable per-iteration
slowdown in training. (As discussed above, each RevNet update is about
1.5-2$\times$ more expensive, depending on the implementation.) We found it
surprising that the performance matched so closely, because reversibility would
appear to be a significant constraint on the architecture, and one might expect
large memory savings to come at the expense of classification error.

{\bf Impact of numerical error.} 
As described in Section \ref{gradientcomp}, reconstructing the activations over
many layers causes numerical errors to accumulate. In order to measure the
magnitude of this effect, we computed the angle between the gradients computed
using stored and reconstructed activations over the course of training.
Figure~\ref{fig:auto_vs_manual} shows how this angle evolved over the course of
training for a CIFAR-10 RevNet; while the angle increased during training, it
remained small in magnitude. Figure~\ref{fig:auto_vs_manual} also shows
training curves for CIFAR-10 networks trained using both methods of computing
gradients. Despite the numerical error from reconstructing activations, both
methods performed almost indistinguishably in terms of the training efficiency
and the final performance.

\section{Conclusion and Future Work}
We introduced RevNets, a neural network architecture where the activations
for most layers need not be stored in memory.
We found that RevNets provide considerable gains in memory
efficiency at little or no cost to performance.  As future work, we are
currently working on applying RevNets to the task of semantic
segmentation, the performance of which is limited by a critical memory
bottleneck --- the input image patch needs to be large enough to process high
resolution images; meanwhile, the batch size also needs to be large enough to
perform effective batch normalization (e.g. \citep{zhao2016pyramid}). We also intend
to develop reversible recurrent neural net architectures; this is a particularly
interesting use case, because weight sharing implies that most of the memory cost
is due to storing the activations (rather than parameters). We envision our reversible \rname{} as a
module which will soon enable training larger and more powerful
networks with limited computational resources.

\bibliographystyle{abbrvnat}
\bibliography{main}

\newpage
\section{Appendix}
\subsection{Experiment details}
For our CIFAR-10/100 experiments, we fixed the mini-batch size to be 100. The learning rate was initialized to 0.1 and decayed by a
factor of 10 at 40K and 60K training steps, training for a total of 80K steps. The
weight decay constant was set to $2 \times 10^{-4}$ and the momentum was set to 0.9. We subtracted the mean image, and augmented the dataset with random cropping and random horizontal
flipping.

For our ImageNet experiments, we fixed the mini-batch size to be 256, split across 4
Titan X GPUs with data parallelism \citep{simonyan2014very}. We employed synchronous SGD \citep{chen2016revisiting} with momentum of 0.9. The model was trained for 600K
steps, with factor-of-10 learning rate decays scheduled at 160K, 320K, and 480K
steps. Weight decay was set to $1 \times 10^{-4}$.  We applied standard input
preprocessing and data augmentation used in training Inception networks
\cite{szegedy2015inception}: pixel intensity rescaled to within [0, 1], random
cropping of size $224 \times 224$ around object bounding boxes, random scaling, random horizontal
flipping, and color distortion, all of which are available in TensorFlow. For the original ResNet-101, We were
unable to fit a mini-batch size of 256 on 4 GPUs,
so we instead averaged the gradients from two serial runs with mini-batch size
128 (32 per GPU).  For the RevNet, we were able to fit a mini-batch size of 256 on 4 GPUs (i.e.~64 per GPU).

\subsection{Memory savings}
Fully realizing the theoretical gains of RevNets can be a non-trivial task and
require precise low-level GPU memory management. We experimented with two different implementations within TensorFlow:

With the first, we were able to reach
reasonable spatial gains using ``Tensor Handles'' provided by TensorFlow, which
preserve the activations of graph nodes between calls to \texttt{session.run}.
Multiple \texttt{session.run} calls ensures that TensorFlow frees up
activations that will not be referenced later. We segment our computation graph
into separate sections and save the bordering activations and gradients into
the persistent Tensor Handles. During the forward pass of the backpropagation
algorithm, each section of the graph is executed sequentially with the input
tensors being reloaded from the previous section and the output tensors being
saved for use in the subsequent section.  We empirically verified the memory
gain by fitting at least twice the number of examples while training ImageNet.
Each GPU can now fit a mini-batch size of 128 images, compared the original
ResNet, which can only fit a mini-batch size of 32. The graph splitting trick
brings only a small computational overhead (around 10\%). 

The second and most significant spatial gains were made by implementing each residual stack as a \texttt{tf.while\_loop} with the \texttt{back\_prop} parameter set to \texttt{False}. This setting ensures that activations of each layer in the residual stack (aside from the last) are discarded from memory immediately after their utility expires. We use the \texttt{tf.while\_loop}s for both the forward and backward passes of the layers, ensuring both efficiently discard activations. Using this implementation we were able to train a 600-layer RevNet on the ImageNet image classification challenge on a single GPU; despite being prohibitively slow to train this demonstrates the potential for massive savings in spatial costs of training extremely deep networks.

\newpage
\subsection{Sample implementation}
\begin{python}
import tensorflow as tf

def forward(inputs, weights, weights_per_layer):
    """Perform forward execution of a reversible stack.
    
    Args:
      inputs: A pair of Tensors to be propagated through
        the stack.
      weights: A TensorArray containing
        `weights_per_layer*layer_count` elements.
      weights_per_layer: An integer.
    Returns:
      outputs: A pair of Tensors with the same shape as
        `inputs`.
    """
    def loop_body(layer_index, inputs, weights):
        layer_weights = weights.gather(
            tf.range(i*weights_per_layer,
                     (i+1)*weights_per_layer))
        outputs = execute_layer(inputs, layer_weights)
        return (layer_index+1, outputs, weights)
    
    _, outputs, _ = tf.while_loop(
        lambda i, *_: i < layer_count,
        loop_body,
        [tf.constant(0), inputs, weights],
        parallel_iterations=1,
        back_prop=False)
    return outputs
\end{python}
\newpage
\begin{python}
def backward(
  outputs, output_grads, weights, weights_per_layer):
    """Perform backpropagation of a reversible stack.
    
    Args:
      outputs: A pair of Tensors, the outputs from a
        reversible stack.
      output_grads: A pair of Tensors, the gradients
        w.r.t. the `outputs`.
      weights: A TensorArray containing
        `weights_per_layer*layer_count`
      weights_grads: A TensorArray containing
        `weights_per_layer*layer_count` elements.
      weights_per_layer: An integer.
    Returns:
      outputs: A pair of Tensors with the same shape as
        `inputs`.
    """
    def loop_body(layer_index,
                  outputs,
                  output_grads,
                  weights,
                  weights_grads):
        layer_weights = weights.gather(tf.range(
          i*weights_per_layer, (i+1)*weights_per_layer))
        (inputs, input_grads,
         layer_weights_grads) = backprop_layer(
            outputs, output_grads, layer_weights)
        weights_grads = weights_grads.scatter(
          tf.range(i*weights_per_layer,
                   (i+1)*weights_per_layer), 
          layer_weights_grads)
        return (layer_index-1,
                inputs,
                input_grads,
                weights, 
                weights_grads)
    
    (_, inputs, input_grads, _,
     weights_grads) = tf.while_loop(
        lambda i, *_: i >= 0,
        loop_body,
        [tf.constant(layer_count - 1), outputs,
         output_grads, weights, weights_grads],
        parallel_iterations=1,
        back_prop=False)
    return inputs, input_grads, weights_grads
\end{python}
\newpage
\begin{python}
def backprop_layer(
  outputs, output_grads, layer_weights):
    """Perform backpropagation of a reversible layer.
    
    Args:
      outputs: A pair of Tensors, the outputs from 
        this reversible layer.
      output_grads: A pair of Tensors, the gradients
        w.r.t. the `outputs`.
      weights: A pair of Tensors, the first holding
        weights for F and the second holding weights
        for G.
    Returns:
      outputs: A  Tensor with the same shape as
      `inputs`.
    """
    # First, reverse the layer to retrieve inputs
    y1, y2 = output[0], output[1]
    F_weights, G_weights = weights[0], weights[1]
    
    # lines 2-4 of Algorithm 1
    z1_stop = tf.stop_gradient(y1)
    
    G_z1 = G(z1_stop, G_weights)
    x2 = y2 - G_z1
    x2_stop = tf.stop_gradient(x2)
    
    F_x2 = F(x2_stop, F_weights)
    x1 = y1 - F_x2
    x1_stop = tf.stop_gradient(x1)
    
    # Second, compute gradients
    y1_grad = output_grads[0]
    y2_grad = output_grads[1]
    
    z1 = x1_stop + F_x2
    y2 = x2_stop + G_z1
    y1 = z1
    
    # lines 5-9 of Algorithm 1
    z1_grad = tf.gradients(
      y2, z1_stop, y2_grad) + y1_grad
    x2_grad = tf.gradients(
      y1, x2_stop, z1_grad) + y2_grad
    x1_grad = z1_grad
    
    F_grads = tf.gradients(y2, G_weights, y2_grad)
    G_grads = tf.gradients(y1, F_weights, z1_grad)
        
    # Finally, construct return values
    inputs = (x1_stop, x2_stop)
    input_grads = (x1_grad, x2_grad)
    weight_grads = (F_grads, G_grads)
        
    return inputs, input_grads, weight_grads
\end{python}

\end{document}